\ifdefined\pdfminorversion\pdfminorversion=4\fi
\RequirePackage[T1]{fontenc}
\documentclass[letterpaper,10pt,conference]{ieeeconf}
\IEEEoverridecommandlockouts
\usepackage{amsmath,amssymb}
\usepackage{booktabs}
\usepackage{graphicx}
\usepackage{xcolor}
\usepackage{cite}
\usepackage{url}

\newsavebox{\notedtablebox}
\newcommand{\tablenote}[1]{%
  \par\vspace{4pt}\noindent
  \begin{minipage}{\wd\notedtablebox}
    \footnotesize\raggedright #1\par
  \end{minipage}\par}

\title{Nine Trials to Recover: A Reproducible Benchmark\\
for Repertoire-Free Soft-Robot Damage Adaptation}

\author{Siyuan Zhang$^{1}$%
\thanks{$^{1}$Siyuan Zhang is with the Robotics Department, University of Michigan,
Ann Arbor, MI, USA.}%
}

\begin{document}
\maketitle

\begin{abstract}
We present a repertoire-free benchmark for soft-robot damage recovery under nine online trials.
The protocol pairs damage masks within each morphology, retains a measured nominal fallback,
and separates method development from evaluation on new bodies. A Gaussian-process
expected-improvement (GP-EI) reference controller adapts actuator phases using three initialization
probes and six feedback-selected rollouts. Across two disjoint 75-body cohorts, it outperforms
random search, Sobol, CEM, and CMA-ES under equal budgets. GP-EI improves the worst-mask gain on
69/75 confirmation bodies and exceeds these four baselines by $0.235$--$0.310$ mean worst-mask reward.
A TuRBO-style local GP is the closest comparator; the paired confidence interval includes zero.
Post-confirmation fixed-controller replay finds 5.06 voxel widths of mean recovery together with
a 0.0031 increase in worst-mask p99 geometric edge strain; a strict no-added-demand deployment
gate retains 48.7\% of mean gain. In a frozen development stress test that physically removes
10\% of occupied voxels, GP-EI improves 71/75 bodies and exceeds official BoTorch TuRBO by
$0.356$ mean worst-mask gain. Together, the replayable controllers,
body-level inference, and frozen-cohort
evaluation provide a reference for measuring the added value of learned damage priors and
future adaptation methods.
\end{abstract}

\section{Introduction}

Robots operating outside a laboratory must cope with physical changes that were not represented
at design time~\cite{bongard2006resilient,cully2015adapt}. A convincing recovery result requires more than a high average reward: it
must reveal the physical-query cost, retain failed trials, prevent favorable damage masks from
inflating the sample size, and separate development from confirmation. We target this evaluation
gap in a low-data setting---new morphology, new actuator failure, no retraining, no
offline repertoire, and only nine complete trials---while retaining the damaged nominal gait as
a deployment fallback.

Prior work has shown that robots can adapt after damage by consulting an offline behavior map and
using Bayesian optimization (BO) to select compensatory behaviors~\cite{cully2015adapt}. Such
repertoires are powerful but require extensive precomputation for a particular platform. Other
approaches exploit differentiable regeneration or learned adaptation~\cite{horibe2022damage}.
We ask: \emph{how far can direct numerical feedback go when the
body is previously unseen, no damage-conditioned repertoire is available, and only nine complete
rollouts can be purchased?}

\begin{figure*}[t]
    \centering
    \includegraphics[width=\textwidth]{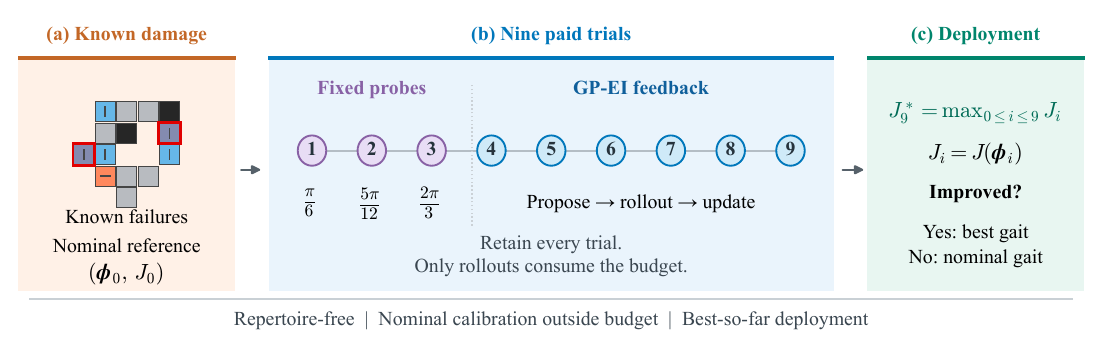}
    \caption{The repertoire-free nine-trial recovery loop. The damaged nominal gait is a shared observation outside the paid budget and
    permanent deployment fallback. Three fixed-radius probes initialize the surrogate; six
    acquisition-selected rollouts close the loop before the best measured gait is deployed.}
    \label{fig:overview}
\end{figure*}

We answer this question with the loop in Fig.~\ref{fig:overview}, implemented in EvoGym~\cite{bhatia2021evogym}. Each controller consists only of the
phases of open-loop sinusoidal actuators. A Gaussian-process (GP) surrogate includes the damaged
nominal controller as a shared observation; three paid probes span fixed local radii, and expected
improvement (EI) selects the remaining six trials from a mixed global/local pool. This construction
provides a compact, replayable reference implementation. It establishes a performance bar for
testing whether a prior, safety layer, or optimizer improves low-budget recovery.

Three design choices support the comparison. First, every method receives the same nine-rollout
budget and nominal fallback, with every exploratory outcome retained. Second, the primary score
is the worse of two damage masks within each body, making morphology the unit of inference.
Third, the main claim is evaluated twice. The second 75-body cohort was generated with
new morphology and damage seeds, had zero overlap with all earlier cohorts, and was sealed before
any damage or optimizer outcome was evaluated.

The benchmark combines an actionable controller baseline with a protocol for testing improvements
on new morphologies. Our contributions are:
\begin{itemize}
    \item a repertoire-free cross-morphology benchmark and evaluation protocol with equal
    nine-rollout budgets, two masks per body, body-level worst-mask inference, unstable-rollout
    retention, and a prospectively sealed confirmation cohort;
    \item a replayable GP-EI reference implementation with a shared nominal incumbent and an
    anytime deployable best-so-far controller;
    \item replicated GP-EI gains over random and CEM in a prospective confirmation, followed by
    an equal-budget Sobol, CMA-ES, and TuRBO-style comparison under a locked extension protocol; and
    \item matched component and execution-noise tests, topology-removal and non-flat-terrain
    stress tests, and a fixed-controller audit that reports recovery beside deformation demand.
\end{itemize}

\section{Related Work}

\emph{Damage recovery.} Self-modeling robots infer changes in their embodiment and synthesize
compensatory behavior~\cite{bongard2006resilient}. T-Resilience combines simulation with
transferability estimates for post-damage search~\cite{koos2013resilience}. Intelligent
Trial-and-Error uses a precomputed behavior map and BO~\cite{cully2015adapt}, drawing on the
repertoire perspective formalized by MAP-Elites~\cite{mouret2015map}. Reset-free Trial-and-Error
extends adaptation to task execution without repeated resets~\cite{chatzilygeroudis2018reset}.
Our benchmark isolates the online recovery stage across unseen morphologies: each optimizer
receives the same nominal gait and nine independently reset trials, with zero repertoire cost.

\emph{Controller adaptation.} BO allocates expensive observations through a
probabilistic surrogate and acquisition function~\cite{jones1998ego,snoek2012bo}, including
gait learning on physical robots~\cite{calandra2016gaits}. Automatic prior selection incorporates
information from simulation or related tasks~\cite{pautrat2018prior}; local BO scales search
through trust regions~\cite{eriksson2019turbo}. We compare these online search ingredients with
uniform random search, Sobol designs~\cite{sobol1967distribution}, CEM~\cite{rubinstein2004cem},
and covariance adaptation~\cite{hansen2001cma} under one shared physical-query budget.
Safety-constrained BO~\cite{sui2015safe,berkenkamp2016safe} addresses feasible exploration;
our incumbent rule addresses which measured controller to deploy.

\emph{Morphology and priors.} Multi-material voxel bodies support rich
shape/control co-design~\cite{cheney2013unshackling,bhatia2021evogym}; SoftZoo broadens the
environment and locomotion tasks~\cite{wang2023softzoo}. Differentiable simulation supports
gradient-based optimization~\cite{hu2020difftaichi}, including morphological regeneration after
damage~\cite{horibe2022damage}. Here the damaged structure is held fixed while actuator phases
adapt. Learned and language-derived skills or value maps provide additional structured
priors~\cite{ahn2022saycan,huang2023voxposer}; the benchmark supplies a common endpoint for
testing their incremental recovery value.

\emph{Reproducible evaluation.} Variability across runs and tasks can obscure algorithmic
progress~\cite{henderson2018matters,agarwal2021precipice}. We combine paired body-level intervals,
explicit rollout accounting, and a prospective cohort split with reproducibility practices
for releasing code and complete experimental records~\cite{pineau2021reproducibility}.

\section{Recovery Setup and Search Methods}

\subsection{Morphology, controller, and damage}

\begin{figure*}[t]
    \centering
    \includegraphics[width=\textwidth]{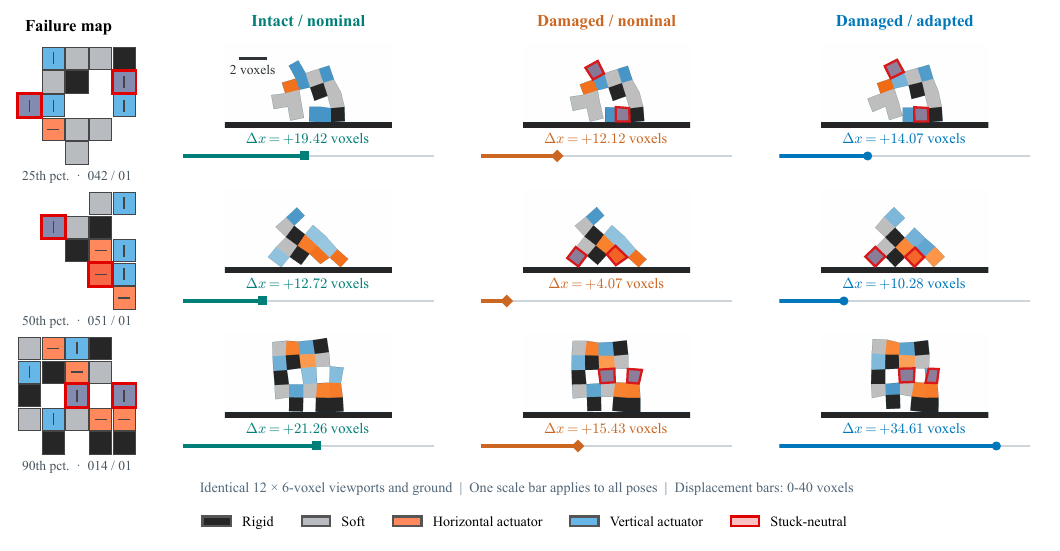}
    \caption{Equal-scale native EvoGym close-ups at step 500. Every pose uses a $12\times6$-voxel
    viewport, identical ground geometry, and horizontal centering; the bars separately show COM
    displacement on a shared 0--40-voxel scale. Red overlays identify stuck-neutral voxels from
    their simulator nodes and preserve the underlying material. Rows are cases nearest the 25th,
    50th, and 90th percentiles of confirmation case-level gain. Each condition is independently
    reset. The supplementary video shows world-frame motion.}
    \label{fig:morphologies}
\end{figure*}

An EvoGym Walker-v0 robot is a $5\!\times\!5$ grid of empty, rigid, soft, horizontal-actuator,
and vertical-actuator voxels. Let $n$ be the number of actuator voxels. The intact controller is
\begin{equation}
u_i(t)=1+0.5\sin(2\pi f t\Delta t+\phi_i),\quad i=1,\ldots,n,
\label{eq:controller}
\end{equation}
where $f$ and the nominal phase vector $\boldsymbol\phi_0$ are fixed for each body. For each damage
case, 20\% of actuators are selected by a frozen seed and forced to neutral, $u_i(t)=1$, while the
voxel remains mechanically present (Fig.~\ref{fig:morphologies}). Damage identities are known;
fault diagnosis is outside our scope.

Let $J(\boldsymbol\phi;d,m)$ denote total reward over a 500-step rollout for morphology $m$ and
damage mask $d$. Evaluating $\boldsymbol\phi_0$ after damage gives $J_0$. After $q$ paid trials,
the deployable score and adaptation gain are
\begin{align}
J_q^\star &= \max\{J_0,J(\boldsymbol\phi_1),\ldots,J(\boldsymbol\phi_q)\},\\
G_q &= J_q^\star-J_0\geq0.
\end{align}
The fallback is a selection rule, not a claim that exploratory rollouts are safe. For inference,
two masks are generated per body and the primary body score is
$S_m=\min_d J_9^\star(d,m)$, preventing correlated masks from inflating the sample size.

\subsection{Nominal incumbent and initialization}

The damaged nominal pair $(\boldsymbol\phi_0,J_0)$ initializes the GP without consuming a paid
query. The first three queries perturb the nominal phase vector with independent componentwise
offsets at radii $\pi/6$, $5\pi/12$, and $2\pi/3$. This fixed schedule supplies conservative,
moderate, and exploratory observations before fitting a highly data-limited surrogate. All phases
are wrapped into $[0,2\pi)$.

\subsection{Surrogate and expected improvement}

Our main implementation uses the periodic kernel
\begin{equation}
k(\boldsymbol\phi,\boldsymbol\phi')=
\exp\!\left[-\frac{2}{\ell^2}\sum_{i=1}^{n}
\sin^2\!\left(\frac{\phi_i-\phi_i'}{2}\right)\right],
\label{eq:kernel}
\end{equation}
with fixed $\ell=\sqrt{\max(1,n/2)}$ and diagonal noise $10^{-6}$. Rewards are standardized using
the observations available at the current query. At each of queries four through nine, EI with
exploration offset $0.01$ is maximized over 2,048 deterministic-seed candidates: half are uniform
on the phase box and half are wrapped perturbations of the current incumbent using the three
initialization radii. This mixed pool keeps global escape routes while concentrating half of the
computation near the best observed controller.

Let $X_q$ and $\mathbf y_q$ contain the nominal observation and the paid
observations available before the next query. With $K_q=k(X_q,X_q)+10^{-6}I$, the standardized
posterior is
\begin{align}
\mu_q(\boldsymbol\phi)&=\mathbf k_q(\boldsymbol\phi)^\top K_q^{-1}\mathbf y_q,\\
\sigma_q^2(\boldsymbol\phi)&=1-\mathbf k_q(\boldsymbol\phi)^\top
K_q^{-1}\mathbf k_q(\boldsymbol\phi).
\end{align}
Writing $z=(\mu_q-y_q^{\max}-\xi)/\sigma_q$ with $\xi=0.01$, the acquisition is
$\mathrm{EI}_q=(\mu_q-y_q^{\max}-\xi)\Phi(z)+\sigma_q\varphi(z)$. Cholesky solves are used
throughout. Fixed hyperparameters keep the small-sample fit and candidate selection replayable.

\begin{center}
\fbox{\begin{minipage}{.92\columnwidth}
\small
\textbf{Algorithm 1: Nine-trial GP damage adaptation}

\textbf{Input:} damaged body $m,d$, nominal phases $\boldsymbol\phi_0$, frequency $f$.

1) Measure $J_0=J(\boldsymbol\phi_0;d,m)$ and set the incumbent to
$(\boldsymbol\phi_0,J_0)$ without charging a search query.

2) For $q=1,2,3$, draw one wrapped perturbation about $\boldsymbol\phi_0$ using radius
$\pi/6$, $5\pi/12$, or $2\pi/3$; evaluate it and update the incumbent only if better.

3) For $q=4,\ldots,9$, fit the fixed-kernel GP, construct 1,024 global and 1,024 local
candidates, evaluate the candidate with maximum EI, and update the incumbent.

4) Return the best measured controller, including the nominal fallback.
\end{minipage}}
\end{center}

The same $J_q^\star$ is recorded after every query. Hence the method is anytime: stopping early
returns the best measured controller, including the nominal fallback. Algorithmic computation is
offline with respect to the simulator; only the selected candidate incurs a rollout.

\subsection{Equal-budget baselines}

Table~\ref{tab:methods} summarizes the equal-budget search methods. Random search draws nine phases
uniformly. Scrambled Sobol search provides a low-discrepancy global design~\cite{sobol1967distribution}.
CEM performs three populations of three. Full-covariance CMA-ES starts at $\boldsymbol\phi_0$ with
population four and initial step size $\pi/2$~\cite{hansen2001cma}. A dependency-free TuRBO-1-style
baseline~\cite{eriksson2019turbo} uses three scrambled-Sobol points, then periodic-RBF posterior
sampling inside one trust region. We additionally adapt the official BoTorch v0.15.1 tutorial
implementation~\cite{balandat2020botorch} to the nine-query regime on development bodies. It shares
the exact three TuRBO-style initialization observations, then fits the tutorial's ARD Mat\'ern-$5/2$
GP and makes six sequential Thompson-sampling queries. We retain its trust-region rules, but use
$q=1$ and stop at nine to match the physical-query budget. Every method
retains $J_0$ as the deployment fallback.

\begin{figure}[t]
    \centering
    \includegraphics[width=\columnwidth]{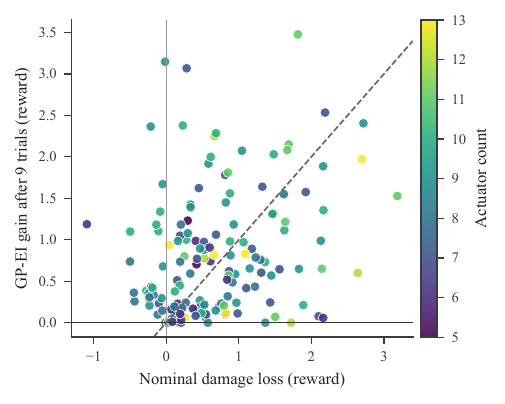}
    \caption{All 150 confirmation cases, without a fitted trend line. The dashed identity line
    denotes full recovery of lost reward; negative damage loss means the damaged nominal
    outscored its intact run. On the minimum-gain mask of each body, independently measured
    mean displacement gain is $0.506$~m (5.06 voxel widths).}
    \label{fig:recoverability}
\end{figure}

\begin{figure*}[t]
    \centering
    \includegraphics[width=\textwidth]{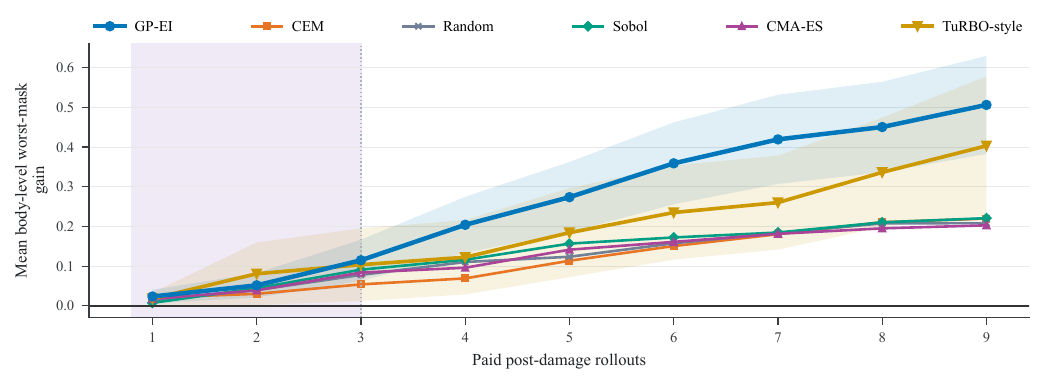}
    \caption{Anytime confirmation performance under identical physical-query budgets. Lines are
    mean body-level worst-mask adaptation gain. Bands, shown only for GP-EI and TuRBO-style
    for readability, are $\pm1.96$ standard errors over 75 bodies. GP-EI separates from
    non-surrogate search after its three initialization probes;
    TuRBO-style is the closest competitor.}
    \label{fig:anytime}
\end{figure*}

\section{Experimental Protocol}

\subsection{Using the benchmark}

\emph{Input and task.} The fixed track is Walker-v0, $5\times5$ bodies, and known 20\%
stuck-neutral damage. Each case provides its voxel matrix, failed actuator IDs, nominal phases
and frequency, and shared measured score $J_0$. The optimizer returns an evaluated phase vector.
\emph{Budget.} Nine additional rollouts of at most 500 steps include initialization,
repeats, and unstable outcomes. The shared $J_0$ calibration and intact characterization are
accounted for separately; fitting and unevaluated candidates consume no physical queries.

\begin{table}[t]
\centering
\vspace*{5pt}
\caption{Nine-query search methods with nominal fallback}
\label{tab:methods}
\setlength{\tabcolsep}{3pt}
\sbox{\notedtablebox}{\begin{tabular}{@{}lll@{}}
\toprule
Method & Initial design & Update / acquisition \\
\midrule
Random & Uniform & None \\
Sobol & Scrambled Sobol & None \\
CEM & Nominal Gaussian & 3 populations $\times$ 3 \\
CMA-ES & Nominal, $\sigma=\pi/2$ & Full covariance, pop. 4 \\
TuRBO-style & 3 Sobol & Local GP sampling \\
BoTorch TuRBO$^*$ & Shared 3 Sobol & Mat\'ern GP + TS \\
GP-EI & 3 fixed radii & Global/local EI \\
\bottomrule
\end{tabular}}
\usebox{\notedtablebox}
\tablenote{TS: Thompson sampling. $^*$Development-only exploratory comparison.}
\end{table}

\emph{Required outputs.} Report paired body-level $S_m$ differences and intervals,
minimum-mask gain curves at every query, coverage, and unstable-rollout rates.
\emph{Reproducibility.} A release package contains cohort JSONs, frozen manifests, per-query
CSVs, reference optimizers, analysis scripts, tests, native replays, checksums, and a usage guide.
It will be made public after review; the evaluation protocol and principal evidence are reported here.
New methods are developed on development bodies; new prospective claims require a fresh sealed
cohort once published confirmation outcomes have been inspected.

\subsection{Cohorts and prospective confirmation}

Table~\ref{tab:cohorts} lists the cohorts. Development contains 75 random morphologies and
150 fixed damage cases. It was used to
identify GP-EI as the strongest online baseline. We then froze a confirmation protocol and built a
new cohort from 150 fixed-seed morphology draws. Eligibility used only intact nominal behavior;
the first 75 unique eligible bodies were retained. The resulting confirmation cohort contains 75
bodies and 150 damage cases, has minimum nominal reward $0.0822$, and has zero morphology overlap
with every prior cohort. Its cohort hash and selection rule were sealed before damage evaluation
or optimizer execution.

All methods use nine queries, 500 simulation steps per query, and method-specific seeds derived
from the experiment seed, anonymous body ID, mask ID, and method label. An unstable simulation is
an observed outcome: it is retained and never resampled. Neither confirmation file has an
incomplete case. The nine-query budget
stays in the single-digit physical-trial regime established by rapid
damage-adaptation demonstrations~\cite{cully2015adapt}, while allocating three initialization
trials and six feedback-selected trials.

\begin{table}[t]
\centering
\vspace*{5pt}
\caption{Evaluation cohorts with morphology- and seed-disjoint confirmation bodies}
\label{tab:cohorts}
\sbox{\notedtablebox}{\begin{tabular}{lrrl}
\toprule
Cohort & Bodies & Cases & Role \\
\midrule
Development & 75 & 150 & Method development \\
Confirmation & 75 & 150 & Frozen main test \\
\bottomrule
\end{tabular}}
\usebox{\notedtablebox}
\tablenote{Two fixed 20\% stuck-neutral masks per body; 500 steps per
rollout; nine paid queries per optimizer; primary unit is the body.}
\end{table}

\subsection{Endpoint and inference}

For each comparison we compute paired differences in body-level worst-mask score at query nine.
Uncertainty is a percentile bootstrap over the 75 bodies with 100,000 draws. One-sided paired
sign-flip tests use one million Monte Carlo draws, followed by Holm correction within each frozen
comparison family. Before observing the original confirmation results, success required both GP--CEM and GP--Random comparisons to have
positive mean and median differences, at least 42/75 body wins, a positive 95\% bootstrap lower
bound, and Holm-adjusted $p<0.05$.

The stronger alternatives were implemented and frozen after the original GP result. Their
parameters were tested only on the independent development cohort, then locked before any Sobol,
CMA-ES, or TuRBO-style outcome on the confirmation cohort. This \emph{confirmation extension}
uses the existing cohort, whose GP outcomes were already known when the extension was designed.

\subsection{Determinism and completeness}

Every stochastic stream is derived by hashing the experiment seed, anonymous body identifier,
mask identifier, and method name. Consequently, adding a method cannot perturb another method's
candidate sequence. The confirmation morphology file and extension runner were sealed before
their respective outcomes. Manifests, hashes, coverage checks, exact replay fields, source,
tests, complete trial logs, and figure-generation code are retained in the release package.
Analysis requires complete cases.

The simulator occasionally terminates unstable trajectories before 500 steps. Such events are
neither deleted nor retried. Their rates among paid confirmation rollouts are 0.30\% for GP-EI,
0.89\% for CEM, 1.11\% for random, 1.04\% for Sobol, 0.81\% for CMA-ES, and 0.52\% for
TuRBO-style. Retention matters because instability is part of the optimizer's realized cost and
can otherwise favor highly exploratory methods.

\emph{Post-confirmation demand audit.} Without making new search queries or
reselecting controllers, we replayed the frozen intact nominal, damaged nominal, and query-nine
GP-EI controller for every confirmation case: once exactly and under five paired phase/frequency
perturbations, totaling 2,700 rollouts. EvoGym represents each voxel by four point masses joined
by side and diagonal springs. From their trajectories we reconstruct each unique structural edge
and define $D_{99}$ as the 99th percentile of
$|\log(\ell_e(t)/\ell_e(0))|$ over edge--time samples. The performance endpoint remains the
minimum adapted-minus-nominal gain across masks; the mechanical endpoint is the maximum
adapted-minus-nominal $D_{99}$ change across masks. This exploratory kinematic demand proxy is
not material stress, fatigue damage, or a safety certificate. We separately replay all nine
already-evaluated GP-EI queries and apply a strict deployment gate: choose the highest-reward
measured controller only if its $D_{99}$ does not exceed damaged nominal. This retrospective
deployment rule adds no search queries and leaves preceding exploration unchanged.

\section{Results}

\subsection{Prospective confirmation}

Fig.~\ref{fig:recoverability} relates nominal damage loss to recovery.
Fig.~\ref{fig:anytime} shows that GP-EI develops a larger anytime recovery gain under the same
trial budget. Table~\ref{tab:confirmation} reports the frozen primary endpoint. GP-EI beats CEM by
$0.2349$ reward on average (95\% CI $[0.0977,0.3645]$) and random search by $0.3105$
($[0.2058,0.4157]$). GP wins 57/75 and 59/75 body-level
comparisons, respectively, and both medians are positive. Both comparisons pass every component
of the pre-registered joint gate. Relative to the damaged nominal controller, GP's worst-mask
gain is positive for 69/75 bodies, with mean $0.5063$ and median $0.2968$. Independent replay of
the selected controllers converts the mean to $0.506$~m, or $5.06$ voxel widths, of additional
forward displacement. Among the 62 primary worst masks with positive nominal damage loss, the
aggregate gain recovers 66.3\% of that loss (median per-body fraction 42.1\%).
The third case in Fig.~\ref{fig:morphologies} reaches 34.61 voxel widths after adaptation versus
21.26 for intact nominal: the nominal gait is a reference, not an optimum. Recovery fractions
can therefore exceed one, or be undefined when nominal loss is zero; unnormalized worst-mask
scores retain these cases in the primary endpoint.

\begin{table}[t]
\centering
\vspace*{5pt}
\caption{Query-nine confirmation results}
\label{tab:confirmation}
\setlength{\tabcolsep}{2pt}
\sbox{\notedtablebox}{\begin{tabular}{@{}lrrrrr@{}}
\toprule
Comparator & Mean & Median & W/L/T & 95\% CI & $p_H$ \\
\midrule
CEM & $0.2349$ & $0.2241$ & 57/16/2 & $[0.0977,0.3645]$ & $0.000376$ \\
Random & $0.3105$ & $0.1954$ & 59/14/2 & $[0.2058,0.4157]$ & $2.0\!\times\!10^{-6}$ \\
Sobol & $0.2577$ & $0.1477$ & 61/13/1 & $[0.1468,0.3689]$ & $1.5\!\times\!10^{-5}$ \\
CMA-ES & $0.2349$ & $0.1517$ & 60/15/0 & $[0.1306,0.3378]$ & $2.2\!\times\!10^{-5}$ \\
TuRBO-style & $0.0995$ & $0.1111$ & 52/23/0 & $[-0.1071,0.2518]$ & $0.1706$ \\
\bottomrule
\end{tabular}}
\usebox{\notedtablebox}
\tablenote{Original confirmation and confirmation extension. Differences: body-level worst-mask GP minus baseline. W/L/T: wins/losses/ties.
CI: body-bootstrap interval. $p_H$: Holm-adjusted one-sided sign-flip test within the
original (first two rows) or extension (last three rows) family.}
\end{table}

\subsection{Cross-cohort and stronger-baseline comparison}

\begin{figure}[t]
    \centering
    \includegraphics[width=\columnwidth]{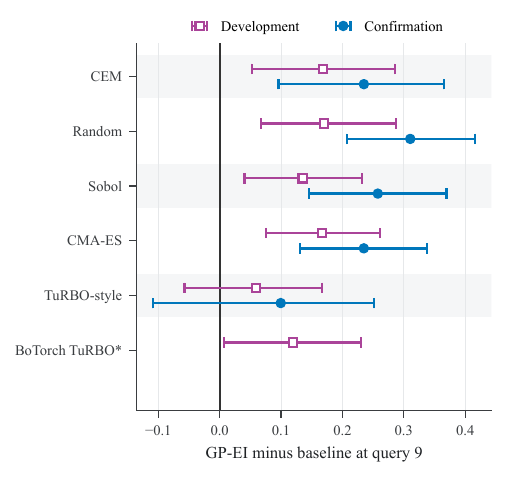}
    \caption{Body-bootstrap mean differences at query 9 on development (purple open squares) and
    confirmation (blue circles). The official BoTorch tutorial adaptation is development-only.
    GP-EI consistently beats non-surrogate search; its TuRBO-style comparison remains unresolved
    on both cohorts.}
    \label{fig:forest}
\end{figure}

The development cohort points in the same direction (Fig.~\ref{fig:forest}): GP exceeds CEM by $0.1679$
(95\% CI $[0.0523,0.2850]$; 55/75 wins) and random search by $0.1701$
($[0.0671,0.2870]$; 52/75 wins). The confirmation cohort uses different morphology, damage,
optimizer, and analysis seeds and was never used to select the method. Directional agreement
therefore supports cross-morphology reproducibility.

On development bodies, GP-EI exceeds Sobol by $0.1352$ (CI $[0.0399,0.2329]$) and CMA-ES by
$0.1667$ ($[0.0753,0.2611]$). These advantages reproduce in the confirmation extension:
$0.2577$ over Sobol and $0.2349$ over CMA-ES, with positive lower confidence bounds and Holm
$p<0.001$. The TuRBO-style local GP is substantially stronger than non-surrogate search. GP-EI's
advantage over it is positive on both cohorts---$0.0588$ in development and $0.0995$ in
confirmation---with intervals spanning zero on both cohorts. The strongest two tested
implementations both use sequential GP-guided search. As a post-confirmation development check,
GP-EI exceeds the official-package TuRBO adaptation by $0.1195$ (CI $[0.0071,0.2307]$;
53/75 wins; one-sided $p=0.019$). The official adaptation minus TuRBO-style is $-0.0607$
($[-0.1265,0.0075]$), with an interval spanning zero. These post-confirmation development
comparisons leave the original confirmatory inference unchanged.

\subsection{Component and execution sensitivity}

We tested whether respecting phase periodicity was itself responsible for the result. A frozen
development ablation replaces Eq.~\eqref{eq:kernel} by an ordinary Euclidean RBF kernel on wrapped
coordinates. It reuses the exact first three phase vectors and scores and consumes the identical
random stream, candidate pools, length scale, EI rule, and nine-query budget. At the same
body-level worst-mask endpoint, periodic minus Euclidean GP is $0.0090$ with CI
$[-0.0851,0.0962]$, median $0$, 37/35/3 wins/losses/ties, and $p=0.426$. It fails the frozen
development gate, so no confirmation experiment was run.

\begin{figure}[t]
    \centering
    \includegraphics[width=\columnwidth]{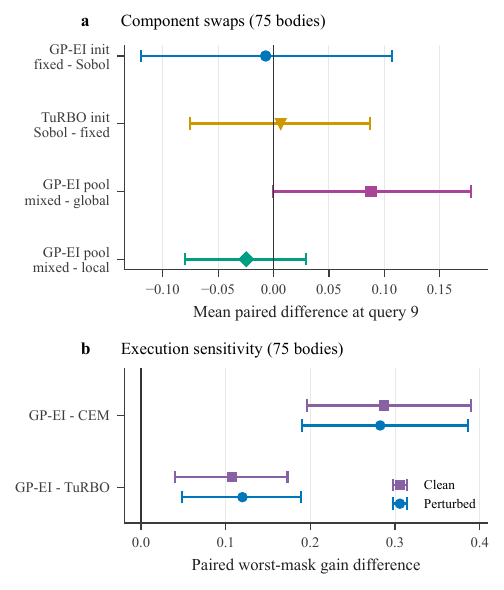}
    \caption{Exploratory development diagnostics. (a) Component swaps on 75 bodies; points are
    mean worst-mask score differences with 95\% body-bootstrap intervals. (b) Paired differences
    in minimum-mask gain on all 75 development bodies, clean (purple squares) and under five paired
    execution perturbations per case (blue), with 95\% body-bootstrap intervals.
    Panel (b) uses gain relative to nominal, not the absolute-score endpoint in Table~\ref{tab:confirmation}.}
    \label{fig:components}
\end{figure}

\begin{figure*}[t]
    \centering
    \includegraphics[width=\textwidth]{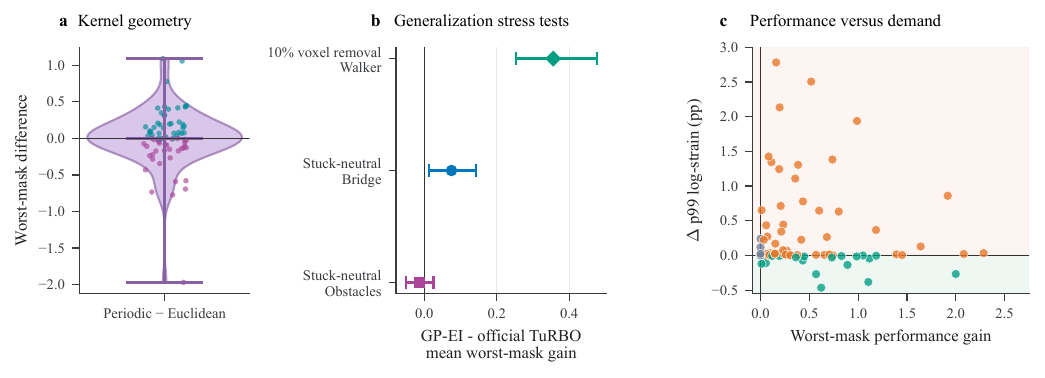}
    \caption{Boundaries of the supported result. (a) The matched periodic-minus-Euclidean kernel
    ablation is centered near zero. (b) Development-only GP-EI differences from official BoTorch
    TuRBO under topology removal and two non-flat tasks (95\% body-bootstrap intervals); green passes
    the frozen multiplicity-corrected gate, blue has a positive interval but misses that gate.
    (c) Confirmation fixed-controller audit: body-level worst-mask
    recovery versus maximum across-mask change in p99 geometric edge strain; green denotes recovery
    without increased demand and orange recovery with increased demand. All panels are
    exploratory rather than original confirmation
    endpoints.}
    \label{fig:robustness}
\end{figure*}

An exploratory development-only component study (Fig.~\ref{fig:components}) then swaps the
initialization and candidate pools. Initialization swaps preserve each optimizer's original
post-initialization random stream. Fixed-radius minus Sobol
initialization for GP is $-0.0071$ (CI $[-0.1197,0.1072]$); Sobol minus fixed-radius initialization
for TuRBO-style is $0.0065$ ($[-0.0752,0.0867]$). GP mixed-pool minus global-only is
$0.0878$ ($[-0.0001,0.1783]$), whereas
mixed minus local-only is $-0.0246$ ($[-0.0800,0.0294]$). All component-swap intervals include
zero, leaving their individual contributions unresolved. The broader baseline comparisons
support surrogate-guided search as a whole; these swaps isolate neither a causal uncertainty
mechanism nor equivalence between GP-EI and TuRBO-style.
Fig.~\ref{fig:robustness} summarizes the kernel ablation, generalization tests, and
performance--demand audit.

\emph{Execution perturbations.} We extended the original 30-body
diagnostic to all 75 development bodies, retaining the same five paired perturbations per case
($0.05$ rad phase SD; $0.02$ log-frequency SD). The 3,000 rollouts evaluate selected controls
and nominal references; they add no search queries or controller reselection. After averaging
repetitions within a case, we subtract its perturbed nominal score and take the minimum gain
across masks. GP-EI retains mean gain $0.500$, versus $0.380$ for TuRBO-style and $0.217$ for CEM.
The paired advantages are $0.120$ (CI $[0.049,0.189]$; 55/75 wins) and $0.283$
($[0.190,0.386]$; 63/75), respectively. The extension followed observation of the initial
subset; on the additional 45 bodies alone, GP--TuRBO is $0.073$ ($[-0.029,0.170]$).

The matching clean minimum-gain comparison is already $0.108$ ($[0.040,0.173]$).
Perturbations change the GP--TuRBO gap by only $0.012$ ($[-0.006,0.033]$), supporting retained
advantage; evidence for noise-induced amplification remains unresolved. An absolute-score endpoint audit,
matching $S_m$, gives $0.059$ ($[-0.058,0.168]$) under perturbation and remains unresolved.
These fixed-controller replays characterize deployment sensitivity under a descriptive gain
endpoint; online acquisition itself remains unperturbed.

\emph{Topology and terrain.} Before outcomes, we froze two connected boundary-patch
removals per development body, each deleting 10\% of occupied voxels while preserving connectivity
and at least four actuators. GP-EI obtains mean worst-mask gain $0.635$ and improves 71/75 bodies;
official BoTorch TuRBO obtains $0.279$. Their paired difference is $0.356$ (CI $[0.253,0.476]$;
62/75 wins; Holm $p<10^{-5}$). CEM and Sobol obtain $0.314$ and $0.263$.
On BridgeWalker, GP-EI also has the largest gain ($0.224$) and exceeds official TuRBO by $0.075$
($[0.013,0.143]$), but misses the six-test multiplicity gate (Holm $p=0.074$). On
ObstacleTraverser, it improves 63/75 bodies yet is not best: GP-EI--official TuRBO is $-0.013$
($[-0.052,0.025]$), while Sobol reaches $0.183$. These are development-only extensions. Because
the frozen two-task gate failed, we did not run the proposed joint terrain--removal experiment.

\emph{Performance and demand.} Exact replay reproduces the frozen rewards to
$4.44\times10^{-16}$. GP-EI adds mean worst-mask reward $0.5063$ (0.506 m, 5.06 voxel widths) and
recovers 66.3\% of aggregate positive nominal loss, while the body-level mechanical endpoint
increases by $0.00309$ (95\% CI $[0.00174,0.00461]$). This is about 0.31 percentage points of
edge stretch and 1.25\% relative to the mean damaged-nominal $D_{99}=0.2475$ on masks selected
by that endpoint. Of 69 recovering bodies, 19 show no increase and 50 show an increase; six bodies
do not recover, split evenly by demand direction. The estimated rank correlation between
performance and demand changes is weak ($\rho=-0.10$, $p=0.39$). Under paired execution perturbations, mean gain
is $0.4640$ and demand change $0.00266$, preserving the same qualitative tradeoff. Adaptation thus
usually restores motion at a small but measurable geometric-deformation cost; whether that cost
accelerates physical damage requires a material-aware model or sensing experiment. The strict
deployment gate retains mean worst-mask gain $0.2468$ (CI $[0.1601,0.3447]$), or 48.7\% of the
standard mean gain, and improves both masks on 39/75 bodies. An improving non-increasing-demand
candidate exists in 107/150 cases, yet 656/1,350 tried candidates exceed nominal demand. Thus a
deployment gate exposes a favorable subset but cannot replace genuinely constrained acquisition.

\subsection{Query allocation and computational cost}

At query three, mean worst-mask gains are $0.1149$ for GP-EI and $0.1037$ for TuRBO-style;
by query six they reach $0.3590$ and $0.2352$, compared with $0.1507$--$0.1723$ for the
non-surrogate methods. The main advantage develops after initialization, when measured rewards
begin to determine acquisition choices (Fig.~\ref{fig:anytime}).

Improvement also spans a broad set of morphologies. At query nine,
GP-EI improves 69/75 bodies and TuRBO-style improves 55/75, whereas CEM, random, Sobol, and CMA-ES
improve 32, 27, 34, and 35 bodies. A positive body gain requires improvement under both masks.

At paid query $q\leq9$, exact GP fitting costs $O(q^3)$ and acquisition scoring costs
$O(Mq^2)$ after the kernel has been formed, with $M=2{,}048$ candidates. These costs are negligible
relative to a 500-step soft-body rollout in our implementation. Memory is $O(Mq+q^2)$; it does not
grow with an offline repertoire. CMA-ES stores an $n\times n$ covariance, whereas the
TuRBO-style implementation uses 256 random Fourier features to avoid a dense
$1{,}024\times1{,}024$ candidate covariance. Computation is CPU-only, with no external model calls.

\section{Discussion and Conclusion}

Nine-trial GP-EI provides a reproducible reference for low-budget recovery across unseen soft-robot
morphologies. Gains over four non-surrogate baselines replicate across disjoint cohorts; the
primary comparison with TuRBO-style remains unresolved. The shared nominal fallback, body-level
inference, and replayable trials make incremental improvements measurable under a common budget.
Development tests identify useful next questions: recovery transfers strongly under topology
removal, whereas optimizer rankings depend on terrain. Morphology-conditioned priors and prior
selection~\cite{pautrat2018prior} can be assessed by their added recovery on fresh bodies under
the same budget.

The evidence covers open-loop controllers, known static faults, and planar EvoGym locomotion
with the reported nine-query configurations. Execution perturbations test deployment sensitivity
of fixed controllers. The performance--demand audit quantifies recovery alongside geometric
deformation and identifies a lower-demand deployment subset. Safe exploration requires
constraints during acquisition and online monitoring~\cite{sui2015safe,berkenkamp2016safe};
a deployment gate acts after trials have occurred. Relating geometric demand to damage
progression requires material-aware models or physical measurements. Contact variability,
fault diagnosis, and hardware transfer remain separate validation steps. The benchmark provides
a common reference for testing these extensions with complete rollout accounting.

\section*{Acknowledgment}
AI-use disclosure: OpenAI Codex assisted with drafting and revising the abstract and Sections I--VI
and with generating experimental, analysis, and plotting code. The authors take responsibility
for the manuscript and associated artifacts.

\bibliographystyle{IEEEtran}
\bibliography{references}

\end{document}